# Jointly Learning Spatial, Angular, and Temporal Information for Enhanced Lane Detection

Muhammad Zeshan. Alam
Department of Computer Science, Brandon, Canada.
(corresponding author e-mail: alamz@brandonu.ca)

*Abstract*—This paper introduces a novel approach for enhanced lane detection by integrating spatial, angular, and temporal information through light field imaging and novel deep learning models. Utilizing lenslet-inspired 2D light field representations and LSTM networks, our method significantly improves lane detection in challenging conditions. We demonstrate the efficacy of this approach with modified CNN architectures, showing superior performance over traditional methods. Our findings suggest this integrated data approach could advance lane detection technologies and inspire new models that leverage these multidimensional insights for autonomous vehicle perception.

## I. INTRODUCTION

Motor incidents caused by human error remain a significant concern, prompting the exploration of innovative solutions for safer transportation [1]. One promising avenue lies in minimizing driver involvement through increased automation, particularly in the domain of perception, as reaction time for autonomous vehicles is not limited by factors such as fatigue [2]. However, developing systems that accurately represent complex real-world scenes is an enormous task.

While an extensive body of research in computer vision exists, further work is required to achieve accurate perception in areas such as lane detection (LD) in challenging environmental conditions. Lane detection systems, pioneered by companies such as Mobileye and Tesla, has made substantial progress with the advent of convolutional neural networks (CNNs) [3], [4].

These CNNs were first used in lane detection problems when combined with RANSAC in [5]. In [6] the precision of lane detection is further improved using ideas such as global information and regular patterns of lanes to provide an anchor-based attention mechanism. VPGNet [7] took a different approach, and combined lane detection with road marking recognition and the vanishing point that guides both tasks for increased performance.

In [8], the joint exploitation of spatial and temporal dimensions in a lane detection network was achieved in three steps; pre-processing CNN-based classification, regression, and lane fitting.Further work included the use of semantic segmentation, such as its combination with optical flow networks in [9], used to derive keyframes in an image. CNNs trained for vision-related tasks continue to evolve, and the decreasing cost of these systems makes their integration into modern vehicles more practical. However, achieving accurate automated perception in novel environments still poses a formidable challenge. For autonomous vehicles to operate beyond controlled environments, they must navigate under varying lighting conditions, weather phenomena, and worn road surfaces.

While deep CNNs have advanced, traditional methods of capturing visual data fall short in representing the complexity of real-world scenes. This limitation compromises the accuracy of information fed into deep learning models.

To bridge this gap, it is imperative to enhance the richness of the visual data used in training learning-based models. While many studies focus on increasing data volume to improve performance, such as the large image dataset ActivityNet [10], richness of data can be achieved through modifications to sensors, optics, or illumination. These adjustments yield more meaningful data, significantly improving lane detection accuracy without requiring these sizeable datasets.

This visual information can be attained through computational imaging, where deep-learning models intersect with light field (LF) imaging. LF imaging captures light rays from varying directions separately at each pixel position [11]–[13], providing not only spatial information but also angular information. This directional information can provide additional discriminatory cues in LD.

Light field acquisition can be done in a many ways including micro-lens arrays (MLAs) [14]–[16], coded masks [17], [18], camera arrays [19], [20], and among these different implementations, MLA-based light field cameras offer a cost-effective approach, leading to commercial LF cameras namely Lytro and Raytrix [14], [15]. We have adopted the first-generation Lytro camera for road lane data acquisition in this work. Inspired by [21]'s approach to utilize LF images in lane detection, this paper introduces a novel approach to lane detection by harmonizing spatial, angular, and temporal data using light field (LF) image sequences captured a regular time intervals. By crafting a lenslet-inspired 2D representation of light fields, we maintain essential spatial and angular information, while the incorporation of temporal dynamics through LSTM networks aims to elevate detection accuracy

## II. INTEGRATING TEMPORAL INFORMATION IN LIGHT FIELD LANE DETECTION

Building on the foundation of employing spatial and angular resolution in addressing lane detection in challenging conditions, we introduce a method that leverages the sequence of light fields for capturing not just spatial and angular, but also temporal information, thus significantly enriching the data available for lane detection. However, incorporating all three information formats into a practical and efficient representation that is compatible with existing state-of-the-art deep learning-based LD methods, the following approach is adopted.

### A. Lenslet Inspired 2D Representation of Light Fields

Decoding raw light field images captured with a Lytro camera results in a set of 11 x 11 perspective images, each with a size of 375 x 375 pixels [22]. Initially, we transform these decoded light field images into a lenslet-inspired 2D representation, which preserves both spatial and angular information, as shown in figure 1. This transformation is crucial for our methodology as it serves as the input to the proposed LSTM-based deep learning model. We transform the central perspective image by integrating macro-pixels that aggregate multiple viewpoints of a single scene point. This transformation is meticulously designed to keep the original image's dimensions unchanged, compensating for the inclusion of macro-pixels by excluding a proportional amount of adjacent spatial pixels, as described in detail in [23]. This approach, while preserving the image's original size, allows for a detailed exploration of both spatial and angular information, crucial for enhancing lane detection in challenging conditions. Through strategic adjustment of macro-pixels, which involves careful selection based on their proximity to the central perspective, this representation emphasizes the angular information's role, offering a nuanced balance between spatial and angular data for optimal lane detection performance. In this paper, the macro-pixel size is empirically set to 2 x 2.

### B. Temporal information integration through LSTM

Subsequently, a sequence of these 2D representations, now enriched with spatial and angular data, is fed into the proposed network architecture, as detailed below. The resulting sequence of features is expected to contain both spatial and angular cues. Unlike [21] where the LSTM is used to extract angular information from a sequence of multiple perspectives of the same image, the proposed approach utilizes the LSTM to harness the temporal information embedded within the sequence of light field images. This distinction is pivotal, as it allows the network to understand changes over time, adding a new dimension to the lane detection capabilities.

### C. Network Architecture

The LSTM architecture is integrated with a backbone feature extractor, a typical convolutional neural network, that processes spatial and angular information. In contrast to the approach outlined in [21], where the feature extractor solely handled spatial data, our method ensures that the extractor now also incorporates angular information. The LSTM, positioned subsequent to this extractor, is specifically tuned to process the temporal dynamics within the sequence of light field images. The implementation details are aligned with the proposed LF representation technique, ensuring that each macro-pixel includes adjacent perspective image pixels based on their proximity to the middle perspective. By prioritizing angular information through macro-pixels, we can delve into the significance of angular cues over spatial details under various conditions, especially those challenging for traditional lane detection systems.

## III. TRAINING

In our research, we employed the Alam et al. [21] dataset, expanded to include 3,000 images across 300 sequences, each with 10 light fields from a single road section. This rich dataset supports temporal, spatial, and angular analysis for lane detection. Originally, sequences had 20 images, but our evaluation found that 10 consecutive frames offered a more effective temporal resolution.

We utilized GoogleNet and VGG16 as feature extractors in training two distinct CNN models. These networks were selected for their diverse performance metrics in prediction accuracy on the ImageNet [24] validation dataset.

To adapt these classification networks for predicting road lane coordinates, we modified them for regression tasks. Specifically, in GoogleNet, we swapped out the 'loss3-classifier', 'prob', and 'output' layers with a fully connected layer with 20 outputs to correspond to lane line point coordinates, and added a regression layer. Similarly, in VGG-16, we removed its fully connected, 'prob', and 'output' layers and integrated a fully connected and regression layer.

The experiments demonstrated the network's ability to maintain prediction accuracy across a range of hyper-parameters, with minor differences in outcomes regardless of the learning rate, optimizer, and batch size configurations. Notably, the batch size was identified as the key hyper-parameter influencing performance enhancement across both models. Enhancing the batch size led to better results, with the maximum feasible batch size being 32, constrained by memory capacity.

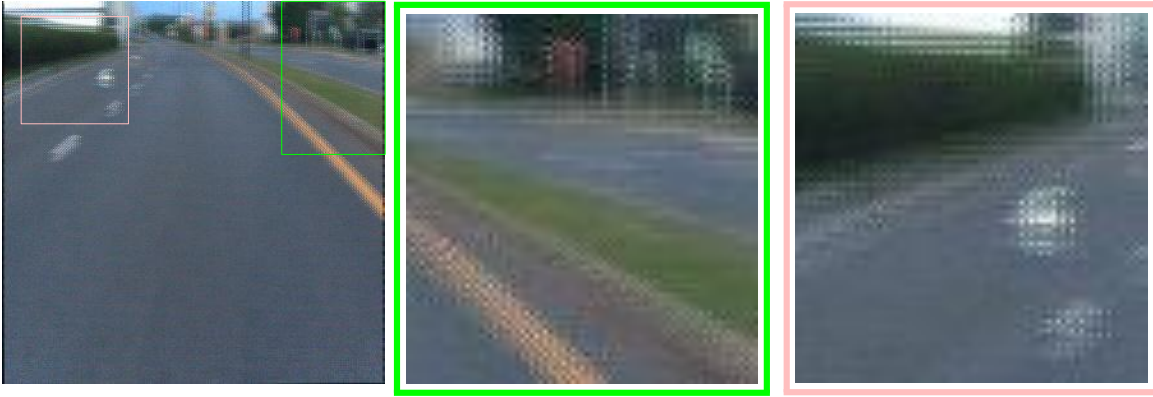

Fig. 1: Sample image of the proposed LF representation with zoomed-in regions focusing on 4 x 4 macro-pixels.

In our chosen configuration, the optimal hyperparameters were a starting learning rate of 3e-4, with a reduction of 0.1 after every 20 epochs. While the optimizer type minimally impacted overall outcomes, the Adam optimizer slightly enhanced predictions, leading to its selection. We fixed batch size at 32 and conducted training over 40 epochs, yielding best results.

In our experiments, the training set consisted of 70% of the total data whereas the test set consisted of just 90 images for both the regular image-based approach and spatial-angular data-based approach, However, the test set for the proposed approach consists of 900 light fields, that is 30% of the total dataset. It should be noted that for the proposed method, the total RMSE is calculated for 90 predictions as each temporal sequence consists of 10 images.

## IV. EXPERIMENTS AND RESULTS

In this section, we provide the evaluation of the proposed multi-dimensional data-based lane detection method. To assess the performance of our regression model, we measure the error using the root mean square error (RMSE), a widely recognized metric. In Figure 2 the proposed LF-based approaches outperform the regular image-based lane detection by a significant margin. Within two different LF approaches, the proposed multi-dimensional information that includes the temporal dimension in addition to spatial and angular information supersedes the method relying only on spatial-angular information.

Numerous deep CNN models tailored for image classification are documented, each affected by various design factors such as the count of trainable parameters, and the architecture's depth and width [25]. To validate our proposed method's effectiveness across diverse CNN frameworks, integral to advanced lane detection techniques and novel feature extraction in learning-based methods, we examined the proposed approach versus standard images and individual LF using another well-known CNN model, namely VGG-16, highlighting the adaptability of our approach to varying network architectures.

The performance gap between the VGG-16 and

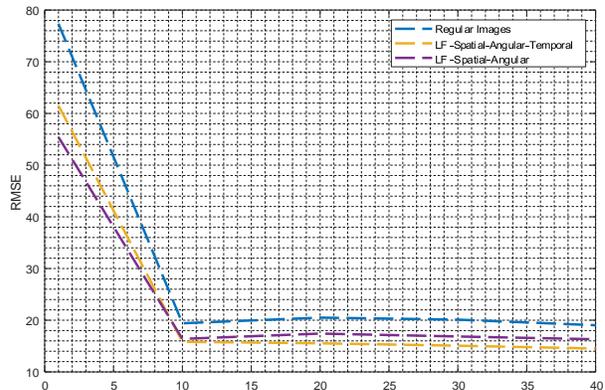

Fig. 2: Comparison of the proposed LD approach with individually selected LF image-based approach [21] as well as regular 2D images-based dataset. GoogleNet is used as a backbone network architecture in the generation of the results

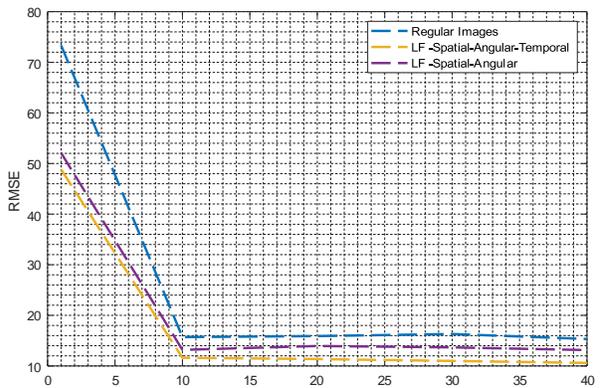

Fig. 3: Comparison of the proposed LD approach with individually selected LF image-based approach [21] as well as regular 2D images-based dataset. Vgg-16 is used as a backbone network architecture in the generation of the results

GoogleNet, as shown in Figure 2 and **??** in marginal. Despite their differing designs, both networks demonstrate that our approach generally surpasses traditional image-based methods and LF-based techniques that process LF images separately. This suggests that our multi-dimensional strategy for road lane detection, which synergizes spatial, angular, and temporal data, can enhance any current lane detection system and inspire new network architectures designed to exploit these combined data types effectively.

## V. CONCLUSION

We propose a multi-dimensional information approach to complement the information missing during lane detection in challenging conditions. Incorporating temporal information alongside spatial and angular information through a novel LSTM-based approach significantly enhances lane detection performance. This methodology not only capitalizes on the detailed lenslet-based 2D representation of light fields but also leverages the sequential nature of these representations to understand temporal changes, offering a comprehensive analysis for improved accuracy in lane detection.

## REFERENCES

[1] J. Boisclair, S. Kelouwani, F. K. Ayevide, A. Amamou, M. Z. Alam, and K. Agbossou, "Attention transfer from human to neural networks for road object detection in winter," *IET Image Processing*, pp. 3544–3556, 2022.

[2] J. Boisclair, S. Kelouwani, A. Amamou, M. Z. Alam, L. Zeghmi, and K. Agbossou, "Image fusion by considering multimodal partial deep neural networks for self driving during winter," in *2021 IEEE Vehicle Power and Propulsion Conference (VPPC)*. IEEE, 2021, pp. 1–6.

[3] T. Ortegon-Sarmiento, S. Kelouwani, M. Z. Alam, A. Uribe-Quevedo, A. Amamou, P. Paderewski-Rodriguez, and F. Gutierrez-Vela, "Analyzing performance effects of neural networks applied to lane recognition under various environmental driving conditions," *World Electric Vehicle Journal*, p. 191, 2022.

[4] M. A. Sajeed, S. Kelouwani, A. Amamou, M. Z. Alam, and K. Agbossou, "Vehicle lane departure estimation on urban roads using gis information," in *2021 IEEE Vehicle Power and Propulsion Conference (VPPC)*. IEEE, 2021, pp. 1–7.

[5] J. Kim and M. Lee, "Robust lane detection based on convolutional neural network and random sample consensus," in *Neural Information Processing: 21st International Conference, ICONIP 2014, Kuching, Malaysia, November 3-6, 2014. Proceedings, Part I 21*. Springer, 2014, pp. 454–461.

[6] L. Tabelini, R. Berriel, T. M. Paixao, C. Badue, A. F. De Souza, and T. Oliveira-Santos, "Keep your eyes on the lane: Real-time attention-guided lane detection," in *Proceedings of the IEEE/CVF conference on computer vision and pattern recognition*, 2021, pp. 294–302.

[7] S. Lee, J. Kim, J. Shin Yoon, S. Shin, O. Bailo, N. Kim, T.-H. Lee, H. Seok Hong, S.-H. Han, and I. So Kweon, "Vpgnet: Vanishing point guided network for lane and road marking detection and recognition," in *Proceedings of the IEEE international conference on computer vision*, 2017, pp. 1947–1955.

[8] Y. Huang, S. Chen, Y. Chen, Z. Jian, and N. Zheng, "Spatial-temporal based lane detection using deep learning," in *Artificial Intelligence Applications and Innovations: 14th IFIP WG 12.5 International Conference, AIAI 2018, Rhodes, Greece, May 25–27, 2018, Proceedings 14*. Springer, 2018, pp. 143–154.

[9] S. Lu, Z. Luo, F. Gao, M. Liu, K. Chang, and C. Piao, "A fast and robust lane detection method based on semantic segmentation and optical flow estimation," *Sensors*, vol. 21, no. 2, p. 400, 2021.

[10] F. Caba Heilbron, V. Escorcia, B. Ghanem, and J. Carlos Niebles, "Activitynet: A large-scale video benchmark for human activity understanding," in *Proceedings of the ieee conference on computer vision and pattern recognition*, 2015, pp. 961–970.

[11] A. Wahab, M. Z. Alam, and B. K. Gunturk, "High dynamic range imaging using a plenoptic camera," in *2017 25th Signal Processing and Communications Applications Conference (SIU)*, 2017, pp. 1–4.

[12] R. Ng, M. Levoy, M. Brédif, G. Duval, M. Horowitz, and P. Hanrahan, "Light field photography with a hand-held plenoptic camera," Ph.D. dissertation, Stanford university, 2005.

[13] M. Z. Alam and B. K. Gunturk, "Hybrid stereo imaging including a light field and a regular camera," in *Signal Processing and Communication Application Conference*, 2016, pp. 1293–1296.

[14] Lytro, Inc., "Lytro - pioneering light field imaging technology," https://www.lytro.com, 2023, accessed: 2024-02-10.

[15] Raytrix GmbH, "Raytrix - 3d light field camera technology," https://www.raytrix.de/, 2023, accessed: 2024-02-10.

[16] M. Z. Alam and B. K. Gunturk, "Hybrid light field imaging for improved spatial resolution and depth range," *Machine Vision and Applications*, vol. 29, no. 1, pp. 11–22, 2018.

[17] S. D. Babacan, R. Ansorge, M. Luessi, P. R. Matarán, R. Molina, and A. K. Katsaggelos, "Compressive light field sensing," *IEEE Transactions on image processing*, vol. 21, no. 12, pp. 4746–4757, 2012.

[18] M. Z. Alam and B. K. Gunturk, "Deconvolution based light field extraction from a single image capture," in *2018 25th IEEE International Conference on Image Processing (ICIP)*, 2018, pp. 420–424.

[19] B. Wilburn, N. Joshi, V. Vaish, E.-V. Talvala, E. Antunez, A. Barth, A. Adams, M. Horowitz, and M. Levoy, "High performance imaging using large camera arrays," in *ACM SIGGRAPH 2005 Papers*, 2005, pp. 765–776.

[20] M. Z. Alam and B. K. Gunturk, "Dynamic range and depth of field extension using camera array," in *2022 30th Signal Processing and Communications Applications Conference (SIU)*, 2022, pp. 1–4.

[21] M. Z. Alam, S. Kelouwani, J. Boisclair, and A. A. Amamou, "Learning light fields for improved lane detection," *IEEE Access*, vol. 11, pp. 271–283, 2023.

[22] M. Z. Alam and B. K. Gunturk, "Light field extraction from a conventional camera," *Signal Processing: Image Communication*, vol. 109, p. 116845, 2022.

[23] M. Z. Alam, S. kelowani, and M. Elsaeidy, "Trade-off between spatial and angular resolution in facial recognition," 2024.

[24] J. Deng, W. Dong, R. Socher, L.-J. Li, K. Li, and L. Fei-Fei, "Imagenet: A large-scale hierarchical image database," in *2009 IEEE conference on computer vision and pattern recognition*. Ieee, 2009, pp. 248–255.

[25] M. Z. Alam, H. F. Ates, T. Baykas, and B. K. Gunturk, "Analysis of deep learning based path loss prediction from satellite images," in *2021 29th signal processing and communications applications conference (SIU)*. IEEE, 2021, pp. 1–4.